\documentclass[11pt,twocolumn]{article}
\usepackage[margin=0.75in]{geometry}
\usepackage{amsmath,amsfonts,amssymb}
\usepackage{graphicx}
\usepackage{float}
\usepackage{adjustbox}
\usepackage{multirow}
\usepackage{booktabs}
\usepackage{pifont}
\usepackage{xcolor,colortbl}
\usepackage{caption}
\usepackage{titlesec}
\usepackage{enumitem}
\usepackage{placeins}
\usepackage{url}
\usepackage{hyperref}

\newcommand{\cmark}{\ding{51}}
\newcommand{\xmark}{\ding{55}}

\definecolor{binary}{HTML}{2196f3}
\definecolor{attribute}{HTML}{4caf50}
\definecolor{motion}{HTML}{9c27b0}

\renewcommand{\arraystretch}{0.9}

\titlespacing{\section}{0pt}{8pt}{4pt}
\titlespacing{\subsection}{0pt}{6pt}{3pt}
\titlespacing{\subsubsection}{0pt}{4pt}{2pt}

\title{Box-QAymo: Box-Referring VQA Dataset for Autonomous Driving}

\author{
Djamahl Etchegaray, Yuxia Fu, Zi Huang, Yadan Luo \\
The University of Queensland, Brisbane, Australia \\
\texttt{\{uqdetche, yuxia.fu, huang, y.luo\}@uq.edu.au}
}

\date{}

\begin{document}

\maketitle

\begin{abstract}
 Interpretable communication is essential for safe and trustworthy autonomous driving, yet current vision-language models (VLMs) often operate under idealized assumptions and struggle to capture user intent in real-world scenarios. Existing driving-oriented VQA datasets are limited to full-scene descriptions or waypoint prediction, preventing the assessment of whether VLMs can respond to localized user-driven queries. We introduce Box-QAymo, a box-referring dataset and benchmark designed to both evaluate and finetune VLMs on spatial and temporal reasoning over user-specified objects. Users express intent by drawing bounding boxes, offering a fast and intuitive interface for focused queries in complex scenes. Specifically, we propose a hierarchical evaluation protocol that begins with binary sanity-check questions to assess basic model capacities, and progresses to (1) attribute prediction for box-referred objects, (2) motion understanding of target instances, and (3) spatiotemporal motion reasoning over inter-object dynamics across frames. To support this, we crowd-sourced fine-grained object classes and visual attributes that reflect the complexity drivers encounter, and extract object trajectories to construct temporally grounded QA pairs. Rigorous quality control through negative sampling, temporal consistency checks, and difficulty-aware balancing guarantee dataset robustness and diversity. Our comprehensive evaluation reveals significant limitations in current VLMs when queried about perception questions, highlighting the gap in achieving real-world performance. This work provides a foundation for developing more robust and interpretable autonomous driving systems that can communicate effectively with users under real-world conditions. Project page and dataset are available at \url{https://djamahl99.github.io/qaymo-pages/}.

\end{abstract}

\section{Introduction}

\begin{figure*}[ht]
    \centering
    \includegraphics[width=\textwidth]{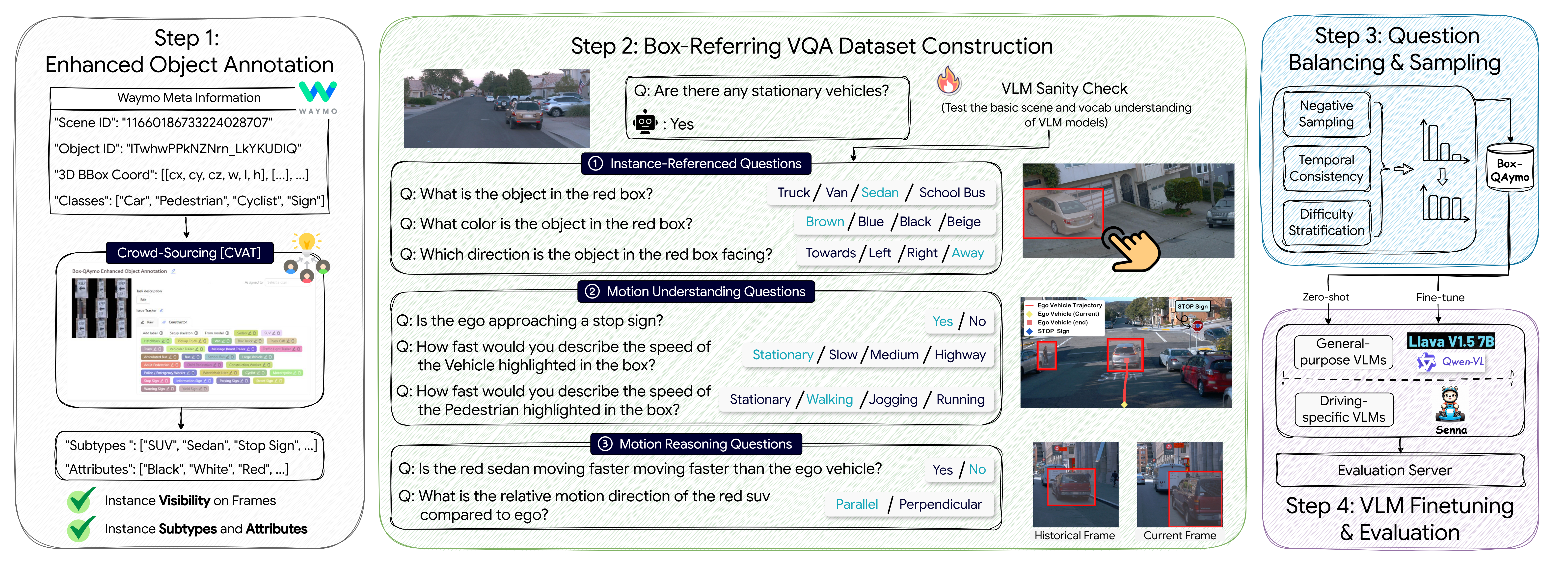}
    \caption{Overview of the Box-QAymo dataset pipeline for evaluating vision-language models (VLMs). \textmd{Step 1 extracts 3D metadata from Waymo which is enhanced with human-annotated semantics. Step 2 introduces box-referenced visual question answering (VQA) tasks spanning instance recognition, motion interpretation, and temporal trajectory reasoning. Step 3 implements rigorous quality control through negative sampling, temporal consistency filtering, and difficulty-aware balancing to ensure a robust and challenging dataset. Step 4 benchmarks general and domain-specific VLMs in zero-shot and fine-tuned settings.}}
    \label{fig:framework}
\end{figure*}

Safety-critical autonomous driving systems depend on effective human-AI communication to foster trust and support widespread adoption. However, recent high-profile incidents involving autonomous vehicles~\cite{national2022summary} have exposed critical gaps in user interaction. Studies show that ~60\% of consumers remain hesitant to adopt fully autonomous vehicles~\cite{aaa2023fear, aaa2025drivers}, with primary concerns reflecting issues with system transparency. These findings underscore the need for frameworks to assess how effectively systems can communicate their understanding and intentions to human users.

Despite this need, progress in evaluating human-AI communication has been limited by the concentration of domain-specific data. Researchers lack access to semantically rich data required to develop and benchmark communication frameworks. Major datasets like Waymo Open~\cite{waymo}, nuScenes~\cite{DBLP:conf/cvpr/CaesarBLVLXKPBB20} and KITTI~\cite {DBLP:conf/cvpr/GeigerLU12} represent millions of dollars in labeling, however, these efforts are constrained by limited class vocabularies. The result is that academic progress in evaluating human-AI communication for autonomous driving lags behind perception capabilities, despite being necessary to enable adoption.

Existing vision-language models adapted for autonomous driving focus predominantly on motion planning while neglecting the equally critical challenge of communicating perceptual understanding to users. Recent works like Drive-LM~\cite{drivevlm} and Talk2Drive~\cite{DBLP:conf/itsc/CuiYZMLLCPW24} demonstrate impressive open-source datasets that integrate LLMs into the planning stage with user intentions. Commercially, Wayve has released LINGO~\cite{DBLP:conf/eccv/MarcuCHKHCNBKSAS24} and Waymo has released EMMA models that can handle many tasks, including 3D object detection, planning, and reasoning. All of these models require substantial data for training and evaluation to effectively encompass many different scenarios. This creates a significant disconnect between research evaluation and real-world deployment scenarios. In practice, autonomous systems must operate with uncertain, incomplete perceptions while simultaneously communicating their confidence and reasoning to users. Current evaluation frameworks fail to capture this dual challenge, leading to models that can plan effectively but cannot explain their limitations or reasoning processes. This limitation becomes particularly problematic when these models encounter edge cases or unusual scenarios in which clear communication of uncertainty becomes critical to maintaining user trust and safety.

The limited semantic diversity in existing autonomous driving datasets constrains the development of models capable of communicating about complex driving scenarios. The vocabularies of popular datasets are restricted to basic objects, such as, vehicles, pedestrians, and cyclists. These vocabulary constraints force models to communicate about complex, nuanced scenarios using overly simplified categorical descriptions. A “vehicle” classification cannot convey the difference between a parked truck, a merging sedan, or an emergency vehicle, distinctions that are crucial for appropriate user communication and trust. This limited object vocabulary directly undermines the development of communication systems that can provide users with the detailed, contextual understanding necessary for informed decision-making in collaborative driving scenarios.

Existing benchmarks either assume unrealistic access to perfect scene understanding or limit evaluation to oversimplified object categories that fail to capture the semantic richness required for effective human communication. To our knowledge, existing datasets do not provide hierarchical evaluation that can assess model understanding across the spectrum from basic perception to complex temporal reasoning about dynamic scenes. This evaluation gap means that researchers cannot adequately measure the progress toward communication systems that would actually be deployable in real-world scenarios. Without benchmarks that reflect true perceptual constraints and semantic complexity, the field lacks the tools necessary to develop reliable, communicative autonomous systems. Addressing this evaluation gap requires datasets that combine realistic perceptual constraints with rich semantic annotations and structured assessment frameworks. 

In this paper, we introduce a novel dataset and evaluation framework (Box-QAymo) that addresses these limitations through crowd-sourced semantic diversity, temporal reasoning assessment, and hierarchical evaluation structures. Our approach expands semantic diversity through crowd-sourced fine-grained class labeling, moving beyond the 3-10 class limitations of existing datasets. We incorporate temporal reasoning evaluation to assess whether multimodal language models can understand object trajectories from visual information alone. Our hierarchical question format provides structured evaluation from binary classification through attribute prediction to complex motion understanding. Our dataset comprises 1,662 binary, 5,403 attribute, and 13,714 motion Q\&A pairs, respectively. By combining semantic richness with temporal understanding and structured evaluation, we can identify specific limitations in current models while providing a pathway for systematic improvement. Our evaluation of general-purpose VLMs and Senna demonstrates the effectiveness of this approach, revealing systematic performance degradation from binary to more complex reasoning tasks, with substantial improvements achievable through targeted finetuning. Our framework establishes a foundation for developing autonomous driving systems that can communicate effectively with users, ultimately supporting the trust and adoption necessary for widespread deployment of autonomous vehicles.

\vspace{-2ex}
\section{Related Work}

Recent efforts in building Visual Question Answering (VQA) datasets for autonomous driving have primarily focused on three core components: perception, prediction, and planning. Perception-focused questions commonly cover environmental elements and conditions like traffic signals \cite{drivelm, reason2drive, drive-with-LLMs, drivevlm, omnidrive, senna}. The questions range from closed-form answers \cite{drivevlm} to free-form scene descriptions \cite{omnidrive, senna}. Prediction-related questions often emphasise the behavioural intentions of surrounding objects to ensure safe path planning for the ego vehicle \cite{drama, reason2drive, drivelm, lingoqa, senna}. Moreover, some datasets incorporate object recognition tasks, involving both static (e.g., type, colour) \cite{drivelm, nuprompt, lingoqa, reason2drive} and dynamic (e.g., speed) \cite{senna, drivelm, drama} attributes, as well as counting and identifying vulnerable or critical objects \cite{drama, drivelm, drivevlm, senna}. Some datasets require reasoning about spatial relationships between objects \cite{nuprompt, reason2drive, drivelm, nuscenesqa, drivegpt4} and ranking their importance \cite{drama, rank2tell}. Planning-related questions can be grouped into meta-actions, justifications, waypoints, attention cues, and user instructions. Meta-actions \cite{senna, drivevlm, omnidrive, drivegpt4, drivelm, lingoqa, drive-with-LLMs} and waypoints \cite{drivevlm} correspond to future ego-vehicle trajectories, while justification \cite{drivegpt4, BDD, BDD-OIA, drivemlm} and attention cues \cite{lingoqa, had, drivelm} are designed to enhance the model’s reasoning and chain-of-thought \cite{drivevlm} capabilities. To facilitate human-AI collaboration and promote safe decision-making, some datasets \cite{had, talk2car, lampilot} introduce user instructions, while others \cite{lingoqa, omnidrive} offer counterfactual reasoning annotations to expose unsafe behaviours.

Although several datasets address either object-centric questions or user-instructed interactions, none have unified both in a way that supports explicit referential grounding. To fill this gap, we propose Box-QAymo, a dataset where users highlight specific objects via bounding boxes and pose instance-grounded, spatiotemporal questions. This design enables a more intuitive form of human-AI interaction and allows for direct evaluation of a model’s referential understanding through box-referenced queries.

\vspace{-2ex}

\section{Dataset Overview}

Our dataset construction methodology follows a three-stage approach as seen in Figure~\ref{fig:framework}: (1) enhanced object annotation using crowd-sourced labeling, (2) box-referring VQA dataset generation based on object characteristics and temporal dynamics, and (3) sampling and balancing strategies for VQA distribution and difficulty control.

\subsection{Enhanced Object Annotation}

\noindent\textbf{Base Dataset Selection.} We build upon the Waymo Open Dataset (WOD) \cite{waymo}, which provides high-quality 3D annotations for autonomous driving scenarios.  We split the WOD validation set evenly into finetuning/validation splits with 101 scenes each. While previous works have opted for nuScenes \cite{nuscenes} for its existing class diversity, we focus on WOD for its superior scene diversity and LiDAR density. While WOD offers robust geometric annotations, its semantic labeling is limited to coarse categories (\textit{i.e.,} pedestrians, cyclists, vehicles, and signs). To enable fine-grained visual question answering, we augment these annotations with detailed semantic labels through crowdsourcing.

\noindent\textbf{Crowdsourced Semantic Labeling.}  Labeling is necessary to enrich objects with enough semantic description for a VLM to single them out. To accomplish this, we employ a crowdsourced annotation strategy to obtain fine-grained semantic labels for approximately 50\% of objects in our dataset, skipping objects with poor visibility. The taxonomy established in Argoverse 2.0 \cite{argoverse} was followed, providing significantly richer semantic categories. For vehicle instances, annotators also provide color labels to enable color-based reasoning questions.

\noindent The annotation pipeline is as follows:
\begin{enumerate}[leftmargin=1em, topsep=0pt, itemsep=2pt, parsep=0pt]
	\item Objects are exported as crops with 3D bounding boxes overlaid for each camera and timestamp they are visible.
	\item We collect up to 9 best crops based on the visibility of the object in each crop.
	\item We fuse these crops as a 3x3 gallery, giving the annotators multiple views to help them with occluded and distant objects.
	\item The object galleries are collected and uploaded to CVAT.
	\item Annotators label instances iteratively and ignore those that are not visible.
\end{enumerate}

\subsection{Box-Referring VQA Dataset Construction}


\begin{table}[h]
	\centering
	\caption{Overview of question subcategories and examples in the Box-QAymo dataset. \textmd{The taxonomy follows a hierarchical structure, starting from basic binary VLM sanity checks and extending to fine-grained object classification and complex spatiotemporal reasoning tasks. This design enables systematic evaluation of VLM capabilities across varying levels of visual understanding.}}
	\label{tab:vlm_questions}
	\adjustbox{width=\columnwidth,center}{%
            \renewcommand{\arraystretch}{1.5}
		\begin{tabular}{ll}
			\hline
			\textbf{Subcategory}                                & \textbf{Example Questions}                                                \\
			\hline
			\rowcolor{binary!20} Movement Status                & "Are there any stationary vehicles?"                                      \\
			\hline
			\rowcolor{binary!20} Orientation                    & "Are there any vehicles moving towards the camera?"                       \\
			\hline
			\rowcolor{attribute!20} Fine-grained Classification & "What type of object is in the red box?"                                  \\
			\hline
			\rowcolor{attribute!20} Color Recognition           & "What color is the object highlighted in red?"                            \\
			\hline
                \rowcolor{attribute!20} Facing Direction           & "What direction is the object in the red box facing?"                            \\
			\hline
			\rowcolor{motion!20} Speed Assessment               & "How fast is the blue sedan moving?"                                      \\
			\hline
                \rowcolor{motion!20} Movement Direction           & "What direction is the object in the red box moving?"                            \\
			\hline
                \rowcolor{motion!20} Relative Motion Analysis       & "Is the green pickup truck traveling faster than the ego vehicle?"        \\
			\hline
			\rowcolor{motion!20} Traffic Element Recognition    & "Is the ego vehicle approaching a stop sign?"                             \\
			\hline
			\rowcolor{motion!20} Trajectory Analysis            & "Are the ego vehicle and the truck on a collision course?"                \\
			\hline
			\rowcolor{motion!20} Relative Motion Direction      & "What is the relative motion direction of the hatchback compared to ego?" \\
			\hline
			\rowcolor{motion!20} Path Conflict Detection        & "Is there a vehicle in the ego vehicle's future path?"                    \\
			\hline
		\end{tabular}
	}
\end{table}

Our question generation follows a principled hierarchical structure, progressing from simple binary queries to complex spatio-temporal reasoning tasks. This design enables systematic evaluation of model capabilities spanning different levels of visual understanding.

\subsubsection{VLM Sanity Check}
We generate binary (yes/no) questions about object presence and basic characteristics, without referring to bounding boxes. These questions are designed to test the common sense and fundamental scene understanding capabilities of VLMs. We use these simple questions as a sanity check to filter out models that lack basic visual-language alignment. Our binary questions include \textcolor{black!50!binary}{\textbf{Movement Status}} and \textcolor{black!50!binary}{\textbf{Orientation}} (see Table~\ref{tab:vlm_questions}).

\subsubsection{Instance-Grounded Questions}
To enable VLMs to answer questions about specific objects, some datasets \cite{drivelm, omnidrive} incorporate object coordinates into the questions. However, VLMs struggle with interpreting precise numerical coordinates, while excelling at visual markers \cite{drivevlm}. Building on evidence that even simple visual prompts, such as red circles, can effectively guide VLM attention\cite{CLIP_circle}, we design instance-grounded questions that use red bounding boxes to highlight target objects. This approach tests whether VLMs can (1) correctly identify the referenced object among multiple instances, (2) classify the fine-grained object types (\textit{e.g.}, sedan vs SUV), and (3) recognize detailed attributes (color, orientation, state). In autonomous driving, this capability is crucial: passengers may need to reference specific vehicles or pedestrians when querying the system's perception. Our instance-grounded questions include \textcolor{black!50!attribute}{\textbf{Fine-grained Classification}} and \textcolor{black!50!attribute}{\textbf{Color Recognition}} (see Table~\ref{tab:vlm_questions}).

Rather than requiring models to infer user intent from ambiguous text alone, bounding-box inputs provide explicit spatial references similar to how humans naturally point to objects when asking questions. This design tests whether models can reason about referred objects or rely on scene-level statistics or textual correlations.

\subsubsection{Motion Reasoning Questions.}
Beyond box-referring comprehension, we investigate whether VLMs can understand motion relationships of referred objects. To address this, we introduce more challenging motion reasoning questions that require understanding object movement over time and spatial relationships between the ego vehicle and surrounding objects. We categorize these as motion understanding questions, divided into implicit and explicit based on their input format.

Motion reasoning questions test VLMs' spatiotemporal understanding through two approaches: implicit questions use single frames to infer motion from visual cues (\textit{e.g.} motion blur, wheel orientation), while explicit questions provide frame sequences for direct temporal analysis. This design reveals whether models truly process temporal information or rely on static scene understanding. Our motion reasoning questions questions include \textcolor{black!50!motion}{\textbf{Speed Assessment}}, \textcolor{black!50!motion}{\textbf{Relative Motion Analysis}}, \textcolor{black!50!motion}{\textbf{Traffic Element Recognition}}, \textcolor{black!50!motion}{\textbf{Trajectory Analysis}} and \textcolor{black!50!motion}{\textbf{Path Conflict Detection}} (see Table~\ref{tab:vlm_questions}).

These questions have answers varying from binary to multiple choice. The multiple choice questions prompt the model to understand speed, heading and movement direction.

\subsection{Question Balancing and Sampling}
\noindent To ensure data set balance and prevent trivial solutions, we implement several sampling strategies.

\noindent\textbf{Negative sampling:} We find samples that do not fit the criteria for the ground truth answer and create additional samples from these criteria - i.e. negatives with varying object types, locations or temporal contexts. In fact, we find for many trajectory questions negative samples are much easier to come by then positive ones, as criteria for determining if a trajectory matches a question are usually quite strict, so finding positives is actually much harder.

\noindent\textbf{Temporal consistency.} We ensure that trajectory-based questions maintain logical consistency across frame sequences through: (1) linearly interpolating object positions to handle missing object information in Waymo, (2) requiring movement patterns to satisfy criteria over multiple consecutive frames to become positive, (3) smooth velocity calculations using temporal derivatives, and (4) selecting timestamps for questions based on temporal convergence patterns rather than instantaneous criteria.

\noindent\textbf{Difficulty stratification.} Questions are categorized by complexity, from basic object detection to complex multi-object trajectory reasoning.  The hierarchical structure allows for granular analysis of model capabilities, from basic object recognition to complex spatio-temporal reasoning in autonomous driving scenarios.

\noindent\textbf{Answer format.} Each multiple-choice question (MCQ) includes multiple-choice answers with 2-4 options, preventing binary guessing while maintaining evaluation precision.

\section{Experiments}

\begin{figure}
    \centering
    \includegraphics[width=\linewidth]{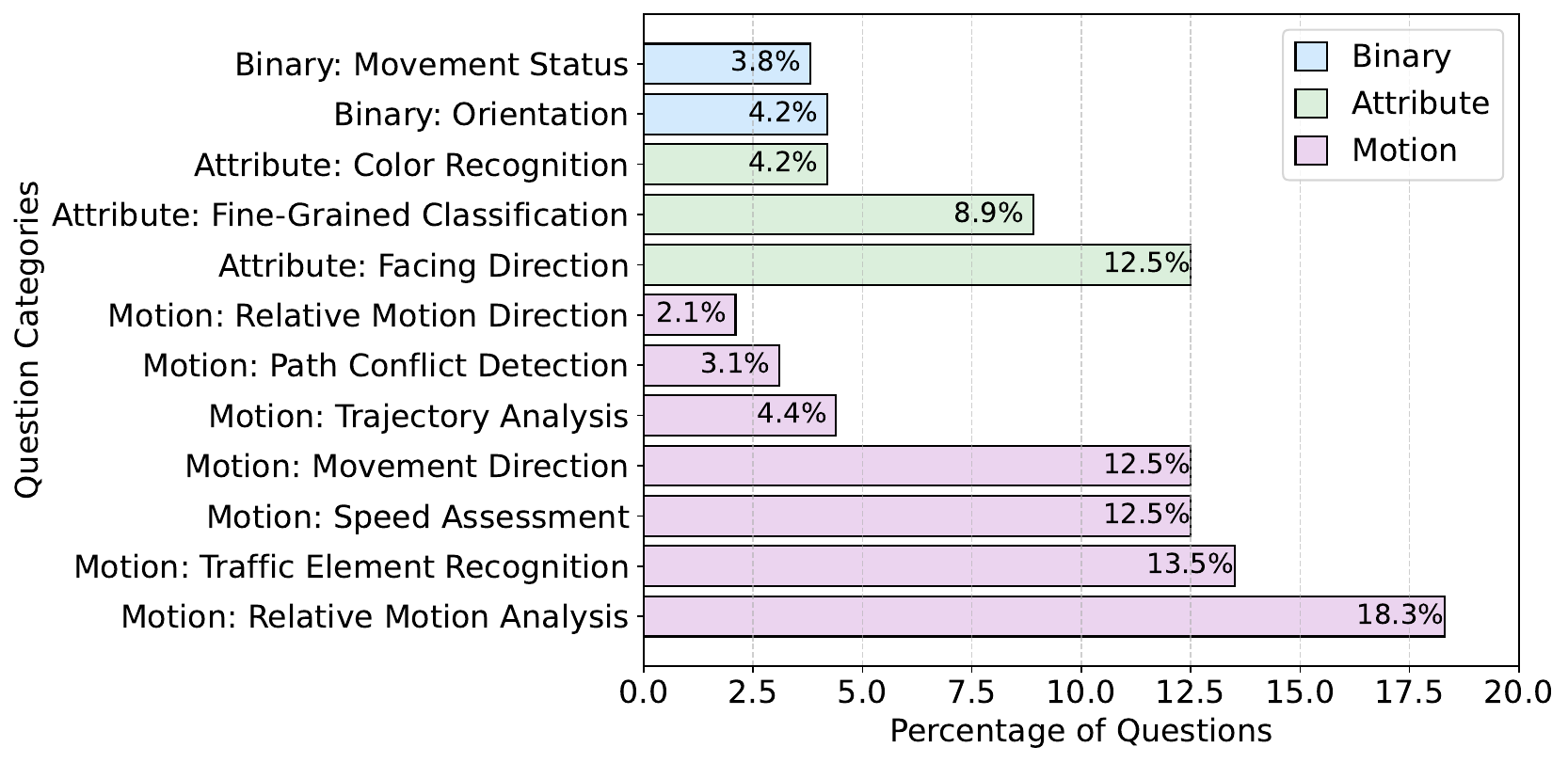}
    \caption{Question distribution across semantic categories in the Box-QAymo dataset. \textmd{The taxonomy spans binary status checks, object attribute recognition, and motion-centric reasoning. Notably, motion-related questions form the majority, emphasizing the dataset’s focus on dynamic scene understanding in autonomous driving scenarios.}}
    \label{fig:question_distribution}
\end{figure}

\subsection{Experimental Setup}

\noindent\textbf{Models Evaluated.} We evaluate three VLMs with varying specialization: \textbf{LLaVA-1.5 7B}~\cite{llava}, a general-purpose multimodal model trained on diverse internet-scale image-text pairs; \textbf{Qwen-7B}~\cite{qwen}, a recently released large-scale vision-language model demonstrating strong zero-shot performance; and \textbf{Senna}~\cite{senna}, a domain-specific model tailored for autonomous driving scenarios, designed to integrate structured driving cues into multimodal reasoning.

\noindent\textbf{Evaluation Protocol.} We evaluated the models in our three-question hierarchies using F1 as the primary metric, supplemented by per-class precision and recall analysis. We report both overall performance and category-specific breakdowns to identify model strengths and limitations.

\noindent\textbf{Steering Prompts.} When running inference, we parse a format prompt to help steer the model to output the desired output format. This could be something like \texttt{"Respond with only the full name of your selection (e.g., brown). Choose the best option out of `red', `yellow', `green' or `brown'."}. This supplies the model with the choices for multiple choice questions, allowing it to give a valid response. 

\noindent\textbf{Response Parsing.} We employ a multistage parser to extract valid answers from free-form model outputs despite the use of steering prompts. This includes: (1) text normalization, (2) exact choice matching, and (3) ranked list handling. Outputs that do not contain a match are marked as invalid.


\noindent\textbf{Finetuning.} We finetuned models with LoRA~\cite{DBLP:journals/corr/abs-2106-09685} for one epoch on our finetuning split using LoRA with rank 128 and alpha 256. We used a learning rate of 2e-4 with cosine scheduling, batch size 8, and bf16 precision. Finetuned models are denoted in tables and figures with \textdagger for single-frame and \textdaggerdbl for two-frame models. The training data was converted to LLaVA chat format and we adapted LLaVA's LoRA finetuning script to suit.

\noindent\textbf{Two-frame Motion.} We investigate whether models can interpret object motion better if we provide the previous camera frame alongside the current, giving the model a chance to spot the differences and interpret the motion of the referred object. The prompt includes both frames to offer temporal context for the referred object.
\vspace{1ex}
\subsection{Overall Performance}
\begin{figure}[!htb]
    \centering
    \includegraphics[width=1.0\linewidth]{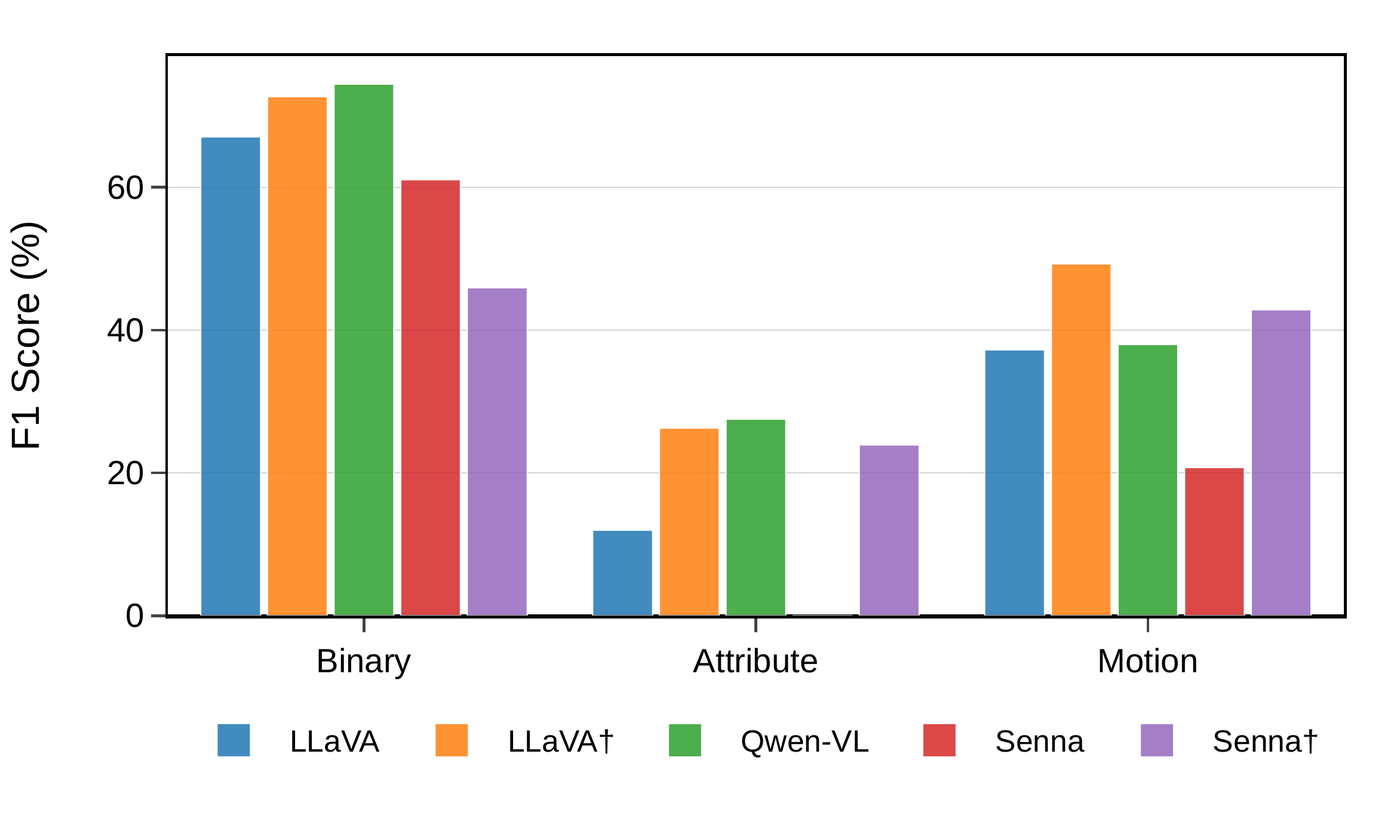}
    \caption{Overall F1 performance across question types for general-purpose and domain-specific VLMs.}
    \label{fig:overall_performance}
\end{figure}

\noindent Figure~\ref{fig:overall_performance} presents the overall performance of all models on the validation set. Finetuned models (\textsuperscript{\textdagger}) outperform their zero-shot counterparts in most cases, with the most significant gains observed in motion-related reasoning, highlighting the importance of task-specific adaptation. Notably, Senna's performance is very low for many tasks and zero for attribute tasks. Qwen-VL is the best of the class, flanked only fine-tuned LLaVA.


\begin{table}[!htb]
\centering
\caption{Performance on binary characteristic questions by subcategory. \textmd{\textdagger indicates single-frame finetuning. \textbf{Bold} indicates best performance, \underline{underline} indicates second-best.}}
\label{tab:binary}
\adjustbox{width=\columnwidth,center}{%
\begin{tabular}{llccc}
\toprule
Question Type & Model & Precision (\%) & Recall (\%) & F1 (\%) \\
\midrule
\multirow{5}{*}{Movement Status} & LLaVA & \underline{76.43} & 62.55 & 68.77\\
 & LLaVA\textsuperscript{\textdagger} & 71.99 & \underline{70.37} & \underline{69.81}\\
 & Qwen-VL & \textbf{76.59} & \textbf{76.13} & \textbf{76.03}\\
 & Senna & 65.23 & 61.11 & 62.20\\
 & Senna\textsuperscript{\textdagger} & 70.43 & 56.17 & 46.91\\
\midrule
\multirow{5}{*}{Orientation} & LLaVA & 72.03 & 59.85 & 65.37\\
 & LLaVA\textsuperscript{\textdagger} & \textbf{77.86} & \textbf{76.02} & \textbf{75.62}\\
 & Qwen-VL & \underline{73.49} & \underline{73.05} & \underline{72.92}\\
 & Senna & 65.65 & 61.90 & 59.94\\
 & Senna\textsuperscript{\textdagger} & 68.66 & 55.02 & 44.96\\
\bottomrule
\end{tabular}
}
\end{table}

\begin{table}[!htb]
\centering
\caption{Performance on attribute questions by subcategory. \textmd{\textdagger indicates single-frame finetuning. \textbf{Bold} indicates best performance, \underline{underline} indicates second-best.}}
\label{tab:attribute}
\adjustbox{width=\columnwidth,center}{%
\begin{tabular}{llccc}
\toprule
Question Type & Model & Precision (\%) & Recall (\%) & F1 (\%) \\
\midrule
\multirow{5}{*}{Color Recognition} & LLaVA & 22.37 & 17.14 & 15.10\\
 & LLaVA\textsuperscript{\textdagger} & \textbf{46.52} & \underline{37.60} & \underline{38.46}\\
 & Qwen-VL & \underline{42.69} & \textbf{50.49} & \textbf{42.49}\\
 & Senna & 0.00 & 0.00 & 0.00\\
 & Senna\textsuperscript{\textdagger} & 39.16 & 29.30 & 28.54\\
\midrule
\multirow{5}{*}{Facing Direction} & LLaVA & \underline{22.98} & 25.05 & 14.49\\
 & LLaVA\textsuperscript{\textdagger} & 9.40 & 24.95 & 13.66\\
 & Qwen-VL & \textbf{30.04} & \textbf{29.64} & \textbf{21.34}\\
 & Senna & 0.00 & 0.00 & 0.00\\
 & Senna\textsuperscript{\textdagger} & 22.26 & \underline{27.08} & \underline{20.72}\\
\midrule
\multirow{5}{*}{Fine-grained Classification} & LLaVA & 11.41 & 5.51 & 6.36\\
 & LLaVA\textsuperscript{\textdagger} & \underline{26.87} & \textbf{29.88} & \textbf{26.73}\\
 & Qwen-VL & 19.60 & \underline{24.93} & 18.81\\
 & Senna & 0.00 & 0.00 & 0.00\\
 & Senna\textsuperscript{\textdagger} & \textbf{27.92} & 21.02 & \underline{22.50}\\
\bottomrule
\end{tabular}
}
\end{table}

\begin{table}[!htb]
\centering
\caption{Performance on motion questions by subcategory. \textmd{\textdagger indicates single-frame finetuning. \textbf{Bold} indicates best performance, \underline{underline} indicates second-best.}}
\label{tab:motion}
\adjustbox{width=\columnwidth,center}{%
\begin{tabular}{llccc}
\toprule
Question Type & Model & Precision (\%) & Recall (\%) & F1 (\%) \\
\midrule
\multirow{5}{*}{Movement Direction} & LLaVA & 24.62 & 25.38 & 19.13\\
 & LLaVA\textsuperscript{\textdagger} & 20.03 & \underline{25.50} & \underline{20.23}\\
 & Qwen-VL & \textbf{26.29} & \textbf{26.80} & 19.68\\
 & Senna & 0.00 & 0.00 & 0.00\\
 & Senna\textsuperscript{\textdagger} & \underline{25.46} & 24.57 & \textbf{22.22}\\
\midrule
\multirow{5}{*}{Path Conflict Detection} & LLaVA & 47.53 & 43.21 & 44.88\\
 & LLaVA\textsuperscript{\textdagger} & 50.06 & 50.02 & 30.28\\
 & Qwen-VL & \underline{57.93} & \textbf{56.63} & \textbf{56.84}\\
 & Senna & \textbf{63.37} & 46.56 & \underline{50.08}\\
 & Senna\textsuperscript{\textdagger} & 53.08 & \underline{51.57} & 33.89\\
\midrule
\multirow{5}{*}{Relative Motion Analysis} & LLaVA & \underline{52.17} & \textbf{54.71} & \underline{48.93}\\
 & LLaVA\textsuperscript{\textdagger} & \textbf{64.57} & \underline{51.80} & \textbf{50.94}\\
 & Qwen-VL & 44.33 & 49.41 & 46.73\\
 & Senna & 49.71 & 28.38 & 35.61\\
 & Senna\textsuperscript{\textdagger} & 44.39 & 50.00 & 47.03\\
\midrule
\multirow{5}{*}{Relative Motion Direction} & LLaVA & 51.48 & 52.51 & \underline{45.12}\\
 & LLaVA\textsuperscript{\textdagger} & \textbf{57.90} & \textbf{61.84} & \textbf{57.40}\\
 & Qwen-VL & \underline{54.56} & \underline{57.29} & 44.16\\
 & Senna & 0.00 & 0.00 & 0.00\\
 & Senna\textsuperscript{\textdagger} & 41.06 & 50.00 & 45.09\\
\midrule
\multirow{5}{*}{Speed Assessment} & LLaVA & 15.53 & 18.69 & 12.92\\
 & LLaVA\textsuperscript{\textdagger} & \textbf{38.76} & \textbf{28.94} & \textbf{27.41}\\
 & Qwen-VL & 16.29 & \underline{22.39} & \underline{15.86}\\
 & Senna & 0.00 & 0.00 & 0.00\\
 & Senna\textsuperscript{\textdagger} & \underline{24.83} & 13.98 & 13.28\\
\midrule
\multirow{5}{*}{Traffic Element Recognition} & LLaVA & 53.81 & 54.56 & 46.07\\
 & LLaVA\textsuperscript{\textdagger} & \textbf{73.11} & \underline{58.33} & \underline{59.73}\\
 & Qwen-VL & \underline{65.39} & \textbf{65.77} & \textbf{65.57}\\
 & Senna & 51.42 & 36.42 & 39.07\\
 & Senna\textsuperscript{\textdagger} & 40.34 & 50.00 & 44.66\\
\midrule
\multirow{5}{*}{Trajectory Analysis} & LLaVA & 44.56 & 43.67 & 43.63\\
 & LLaVA\textsuperscript{\textdagger} & \textbf{98.84} & \textbf{98.97} & \textbf{98.90}\\
 & Qwen-VL & 18.65 & 18.90 & 17.03\\
 & Senna & 35.76 & 15.27 & 20.64\\
 & Senna\textsuperscript{\textdagger} & \underline{92.92} & \underline{95.26} & \underline{93.81}\\
\bottomrule
\end{tabular}
}
\end{table}


\begin{table}[!htb]
\centering
\caption{Single-frame vs two-frame input comparison on motion questions. \textmd{Senna$''$ compares single-frame finetuned Senna to two-frame finetuned Senna. Despite finetuning on temporal data, single-frame input consistently outperforms two-frame input. \textbf{Bold} indicates best performance, \underline{underline} indicates second-best.}}
\label{tab:multiframe_comparison}
\adjustbox{width=\columnwidth,center}{%
\begin{tabular}{lccc}
\toprule
\multirow{2}{*}{Model} & Single Frame & Two Frames \\
 & F1 (\%) & F1 (\%) & Improvement (\%) \\
\midrule
LLaVA & 49.23 & 43.81 & -5.42 \\
LLaVA No Box & 46.94 & \underline{46.33} & -0.61 \\
Qwen-VL & \underline{52.88} & \textbf{50.14} & -2.74 \\
Senna & 20.52 & 21.79 & 1.27 \\
Senna$''$ & \textbf{56.56} & 44.40 & -12.16 \\
\bottomrule
\end{tabular}
}
\end{table}

\begin{table}[!htb]\vspace{2ex}
\centering
\caption{Effectiveness of red bounding boxes at instance-grounding. \textmd{\cmark and \xmark refer to models run with and without bounding boxes, respectively.}}
\label{tab:box_ablation_attribute}
\small
\setlength{\tabcolsep}{4pt}
\begin{tabular}{llcc}
\toprule
                                    &  &  Box: \xmark &  Box: \cmark\\
\multirow{-2}{*}{Question Type} & \multirow{-2}{*}{Model} & F1 (\%) & F1 (\%) \\
\midrule
\multirow{2}{*}{Color Recognition} &  LLaVA & 14.75 & \textbf{15.10}\\
 & Qwen-VL & 40.07 & \textbf{42.49}\\
\midrule
\multirow{2}{*}{Fine-grained Classification} & LLaVA & \textbf{7.64} & 6.36\\
 & Qwen-VL  & 18.65 & \textbf{18.81}\\
\midrule
\multirow{2}{*}{Orientation} & LLaVA & 13.17  & \textbf{14.49} \\
 & Qwen-VL & 19.74 & \textbf{21.34}\\
\bottomrule
\end{tabular}
\end{table}


\subsection{Discussion}



\noindent\textbf{Model Response Quality.} Our evaluation reveals stark differences in models' ability to generate parsable responses, with Senna achieving only 34\% valid responses compared to LLaVA (90\%) and Qwen-VL (99\%). This parsing failure rate represents a fundamental limitation - models that cannot reliably follow structured output formats are unsuitable for safety-critical autonomous driving applications where consistent, interpretable outputs are essential. 

\noindent\textbf{Hierarchy Complexity Validation.} Model evaluation across binary, attribute and motion questions can be found in Table~\ref{tab:binary}, Table~\ref{tab:attribute} and Table~\ref{tab:motion}, respectively. Average F1 scores decrease substantially from binary questions (66.1\% ) to attribute questions (18.3\%) and motion questions (37.6\%).  This strongly validates our complexity assumptions. This pattern reinforces our assumption that binary decisions are the most accessible visual reasoning task, attribute reasoning demands fine-grained visual understanding and semantic knowledge, and motion understanding requires temporal reasoning capabilities that current VLMs lack. Importantly, the intermediate performance on motion tasks (37.6\%) compared to attributes (18.3\%) suggests that the models may rely on static visual cues rather than true temporal understanding.

\noindent\textbf{Senna's Performance.} Senna's poor performance despite being driving-specific reveals the brittleness of narrow task training. The model was optimized for planning (\textit{ e.g.,} given some instructions, respond "go straight") and scene description, making it unable to adapt to structured Q\&A. This highlights the importance of diverse training tasks for robust VLMs. General models (LLaVA, Qwen-VL) show better instruction following and adaptability. There is a significant challenge in task transfer for VLMs in autonomous driving. Development in autonomous driving implies designing domain-specific models like Senna, however, due to a narrowed task focus, these models can fail to transfer to other driving-domain tasks. Models should be trained on diverse tasks within their domain rather than just 1-2 specific task types.

\noindent\textbf{Single vs Two Frames Analysis.} We investigated whether consecutive frames improve motion understanding by comparing single frame vs. two frame input (current + previous frame at 100-ms intervals). Counterintuitively, consecutive frames consistently degrade performance across all models, even when specifically finetuned on temporal data (Senna$''$ : 56.56\% vs 44.40\% F1 in Table~\ref{tab:multiframe_comparison}). This suggests that current VLMs struggle to effectively integrate short-term temporal information for structured reasoning tasks, likely due to training paradigms optimized for single images or longer video sequences rather than frame pairs. 

\noindent\textbf{Box Grounding Effectiveness.} Table~\ref{tab:box_ablation_attribute} shows that box grounding consistently improves Qwen-VL performance in all tasks (average +1.39\% F1), with the largest gains in color recognition (+2.42\% F1). In contrast, LLaVA shows mixed results: improvements in color recognition and orientation (+0.35\% and +1.32\% F1 respectively) but degradation in fine-grained classification (-1.28\% F1). We attribute these differences to architectural design: Qwen-VL's cross-attention mechanism and larger ViT-G/14 encoder enable better separation of visual grounding cues from box artifacts, while LLaVA's simpler MLP projection with ViT-L/14 struggles to filter box-induced noise. These findings highlight that the effectiveness of visual grounding critically depends on model architecture.

\begin{figure}
    \centering
    \includegraphics[width=1\linewidth]{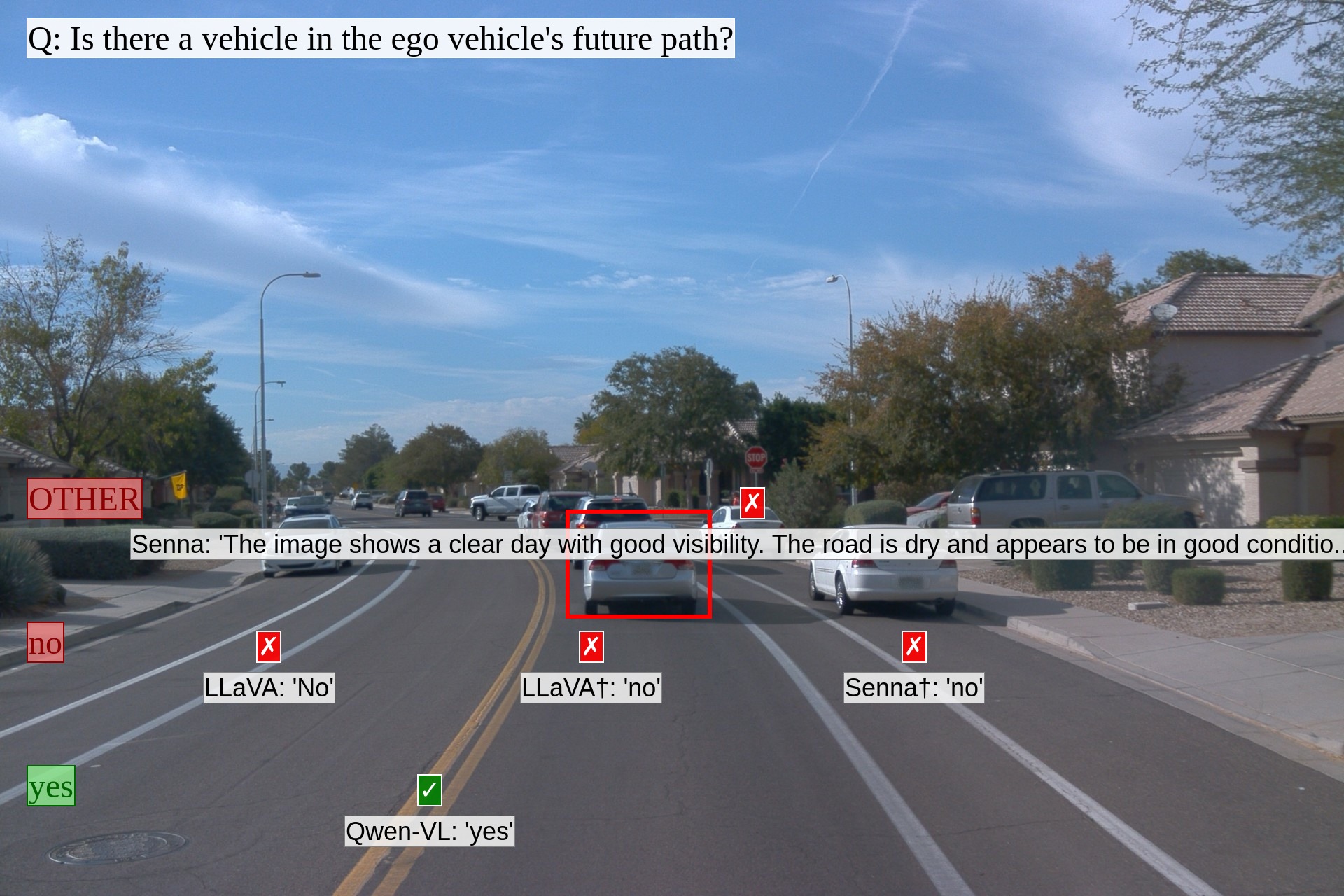}\vspace{1ex}
    \caption{Example of a path conflict detection question and the outputs of pretrained and fine-tuned VLMs. }
    \label{fig:vis}\vspace{2ex}
\end{figure}

\noindent\textbf{Case Study.} Path conflict detection is a critical capability for autonomous vehicles. Figure~\ref{fig:vis} illustrates pretrained and fine-tuned VLM model responses to the given motion query. Despite the clear visual evidence of a leading vehicle in the lane, only Qwen-VL correctly identifies the conflict. LLaVA and its fine-tuned variant fail to recognize the obstruction, responding incorrectly with “no.” The domain-specific VLM Senna also fails to detect the vehicle and instead generates an irrelevant scene-level caption focused on weather and road conditions. This example highlights a key limitation of current VLMs: even with fine-tuning, many models struggle with localized, intent-driven queries requiring precise spatial grounding.


\vspace{-1ex}\section{Conclusion}\vspace{-1ex}
We present the Box-QAymo dataset, to benchmark VLMs under realistic constraints. Our evaluation revealed three critical failures: format adherence, hierarchical performance degradation, and temporal reasoning. These findings challenge current approaches, especially that domain-specific models perform significantly worse on our benchmark. This highlights a critical gap in the demonstration of safety and trust in autonomous driving. When passengers cannot reliably query what an autonomous vehicle perceives, the fundamental promise of AI-enabled driving is not fulfilled. This gap between current capabilities and deployment requirements suggests that achieving trustworthy human-vehicle communication will require rethinking our approach to VLM development, moving beyond task optimization toward architectures that maintain robust reasoning across diverse query types.





\bibliographystyle{plain}
\bibliography{main}


\end{document}